\documentclass[cameraready]{Interspeech}

\usepackage{adjustbox}
\usepackage{multirow}
\usepackage{colortbl}

\newcommand{\pms}[1]{{\scriptsize$\pm$#1}}
\newcommand{\fv}[2]{#1\,/\,#2}

\title{Do Modern Video-LLMs Need to Listen? A Benchmark Audit and Scalable Remedy}

\author[affiliation={1,2}, orcid=0009-0001-6713-3858]{Geewook}{Kim}
\author[affiliation={2}, orcid=0000-0001-6969-4782]{Minjoon}{Seo}

\address{
    $^1$ NAVER Cloud AI, South Korea \\
    $^2$ KAIST AI, South Korea
}

\email{gwkim.rsrch@gmail.com, minjoon@kaist.ac.kr}

\keywords{video large language models, audio-visual question answering, multimodal benchmarks}

\begin{document}

\maketitle

\begin{abstract}
Speech and audio encoders developed over years of community effort are routinely excluded from video understanding pipelines, not because they fail, but because benchmarks never required listening.
We audit 10 video benchmarks and find items largely solvable from visual cues alone: a single-frame probe answers about 76\% of AVQA without audio, suggesting poor measurement of audio-visual reasoning.
Building on LLaVA-OneVision, we attach a speech/audio encoder and compare five compressor architectures under 25-fold token reduction (25 Hz to 1 Hz).
Across 10 benchmarks, with and without filtering, audio yields clear gains on tasks requiring speech comprehension or cross-modal grounding, while vision-centric suites remain largely unaffected.
Our results show that speech encoders play a larger role in video understanding than current benchmarks suggest. We will open-source our work at \url{https://github.com/naver-ai/unimambamia-av}.
\end{abstract}

\section{Introduction}
\label{sec:intro}

Speech and audio encoders such as Whisper~\cite{whisper} are now strong at recognition, speaker analysis, and sound event detection.
Yet these encoders are routinely excluded from video understanding pipelines.
The LLaVA family~\cite{liu2023llava,zhang2024llavanextvideo,li2025llavaonevision,zhang2025llavavideo} and the Qwen-VL family~\cite{qwen2vl} default to ``Video (w/o Audio) $\rightarrow$ Text,'' treating the audio stream as dispensable.
When a user asks a model to summarize a lecture or recap a meeting, understanding naturally requires listening---yet current pipelines silently discard the soundtrack.

Why are capable speech encoders left out?
A structural issue lies in the benchmarks: suites like ActivityNetQA~\cite{actqa}, NExTQA~\cite{xiao2021next}, and TempCompass~\cite{liu-etal-2024-tempcompass-compress} probe visual recognition and temporal structure without requiring audio to answer.
Because the field's primary benchmarks do not require listening, most research has focused on vision-only pipelines, normalizing the omission of audio.

More critically, even benchmarks explicitly designed for audio-visual QA suffer from severe visual shortcuts.
We apply a single-frame shortcut test: feed only the temporally central frame---no audio, no other frames---to GPT-4o and check whether the answer is correct.
The results are striking: \textbf{from a single muted frame, GPT-4o answers 76\% of AVQA~\cite{yang2022avqa} correctly} (Fig.~\ref{fig:filter_rates}).
This suggests that widely used benchmarks---even those marketed as audio-visual---are poorly suited to measuring cross-modal reasoning.
Consider the AVSpeakerBench item in Fig.~\ref{fig:compressor}: the question ``who speaks most quietly'' requires \emph{listening}---a capability speech encoders readily provide but Video-LLMs currently discard.
While recent suites~\cite{sung2024avhbench,worldsense,avspeakerbench,avut} more tightly couple audio-visual cues, legacy suites remain the de facto standard and have not been audited for such shortcuts.

Our experiments show that speech encoders clearly help, especially on the audio-demanding items our audit retains. The real challenge is integrating them \emph{efficiently}.
Audio front-ends produce tokens at 25--50\,Hz; at 25\,Hz, a one-hour video already yields ${\sim}$90K audio tokens, quickly saturating context budgets.
Benchmarks such as VideoMME~\cite{fu2024video} and LongVideoBench~\cite{wu2024longvideobench} routinely include hour-long videos, making this token cost a concrete bottleneck rather than a hypothetical one.
Models that do ingest audio, such as Qwen2.5-Omni~\cite{xu2025qwen25omnitechnicalreport}, feed these tokens uncompressed at 25\,Hz, incurring substantially higher latency (4.1\,s vs.\ 1.0\,s per sample).
Without compressing audio to roughly 1\,Hz, hour-long audio-visual understanding remains impractical.
Starting from LLaVA-OneVision~\cite{li2025llavaonevision}, we attach a speech/audio encoder and compare input strategies (concatenation vs.\ time-aligned interleaving) across 10 benchmarks, with and without single-frame filtering.
We compare five compressor architectures and show that a causal Mamba-based design reduces audio token rates by $25\times$ (from 25\,Hz to 1\,Hz), enabling scalable audio-visual inference for long-form videos.
Time-aligned interleaving with a causal compressor is also the only configuration compatible with streaming inference, where audio arrives incrementally alongside video frames---a practical requirement for real-time applications.
Our contributions are as follows:
\begin{itemize}
    \item We audit 10 video benchmarks with a single-frame filtering protocol, revealing that widely used suites---even those marketed as audio-visual---largely admit vision-only shortcuts. We release the filtered evaluation splits to support fairer assessment of audio-visual models.
    \item We conduct a controlled comparison of input strategies (Table~\ref{tab:ablation}, top) and show that, after single-frame filtering (Table~\ref{tab:ablation}, bottom), audio produces clear gains on tasks requiring speech comprehension or cross-modal grounding---confirming these gains are not artifacts of visual shortcuts. Among five compressor architectures compared under $25\times$ compression, a causal Mamba-based design proves the most stable (Table~\ref{tab:compressor}).
\end{itemize}

\section{Related work}
\label{sec:related}

\paragraph*{Video-LLMs and Benchmarks.}
The LLaVA series~\cite{liu2023llava,li2025llavaonevision,zhang2025llavavideo} and Qwen-VL family~\cite{qwen2vl,xu2025qwen25omnitechnicalreport} are representative examples of multimodal LLMs that typically evaluate on muted video. Audio-visual variants such as VideoLLaMA2~\cite{cheng2024videollama2advancingspatialtemporal} rely on learned-query resampling~\cite{blip2}.
Even dedicated audio-visual suites (AVQA~\cite{yang2022avqa}, Music-AVQA~\cite{Li2022Learning}) admit single-frame shortcuts (Sec.~\ref{subsec:audit}).
Recent benchmarks (AV-Odyssey~\cite{gong2024avodyssey}, OmniBench~\cite{li2024omnibench}, SAVVY~\cite{chen2025savvy}, AVSpeakerBench~\cite{avspeakerbench}, WorldSense~\cite{worldsense}) more tightly couple audio-visual cues. We evaluate on a broad set spanning legacy and recent suites.

\paragraph*{Mamba-based Token Compression.}
Recent vision-language work has moved from sparse frame sampling toward keeping more frames and compressing their tokens, using aggregation (LongVU~\cite{shen2025longvu}, VAMBA~\cite{vamba}), learned-query resampling~\cite{blip2}, or state-space models~\cite{videomamba}. Vision-token compression is now well studied~\cite{blip2,shen2025longvu,vamba,kim2026statespace}; however, \emph{audio}-token compression has not been systematically compared despite 25--50~Hz front-ends~\cite{chu2024qwen2audiotechnicalreport} that produce token counts rivaling or exceeding those of vision.
We present the first systematic comparison of compressor architectures for audio tokens and ask whether designs proven effective for 2-D visual tokens transfer to the structurally different---1-D and inherently causal---audio stream.

\section{Method}
\label{sec:method}

We address two questions: (1)~do current benchmarks genuinely require listening? and (2)~how can audio be integrated efficiently into modern Video-LLMs? We first describe our benchmark audit protocol and then our modeling approach.

\begin{figure}[t]
    \centering
    \includegraphics[width=\linewidth]{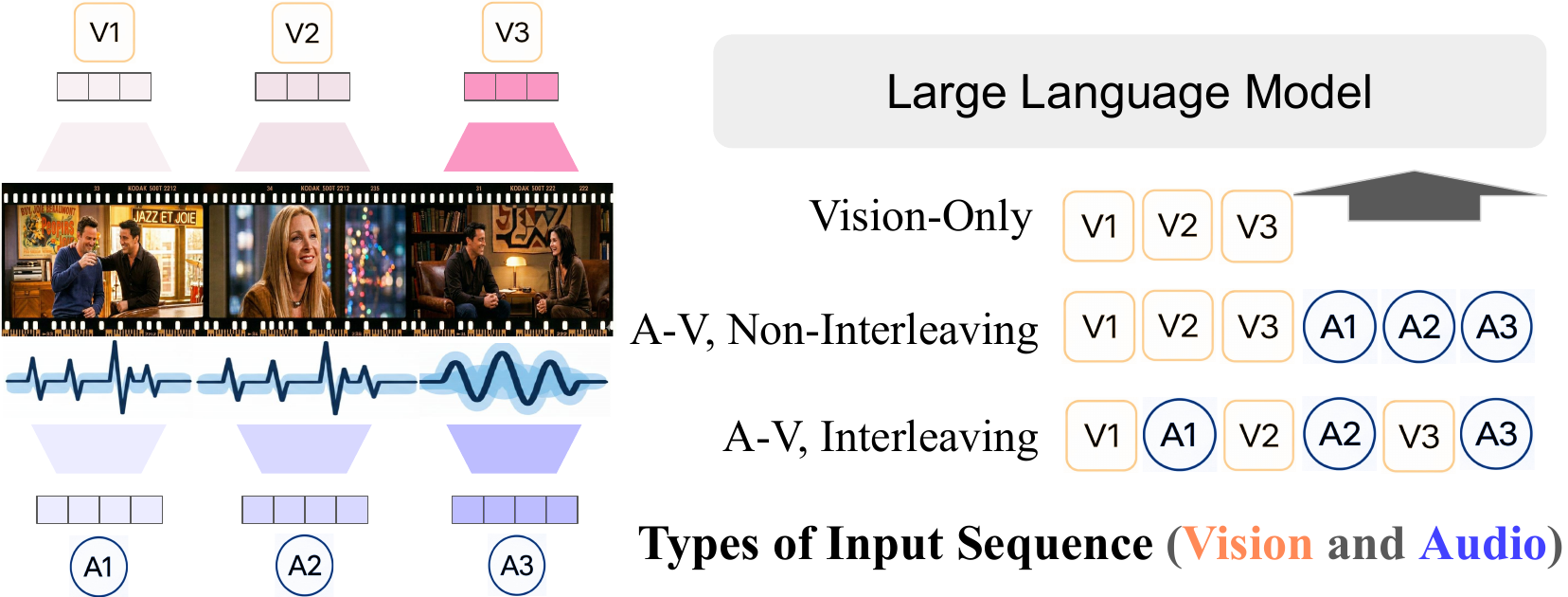}
    \caption{Three input policies for feeding vision and audio tokens to the LLM. Audio tokens may be compressed via Fig.~\ref{fig:compressor}.}
    \label{fig:overview}
\end{figure}

\begin{figure}[t]
    \centering
    \includegraphics[width=\linewidth]{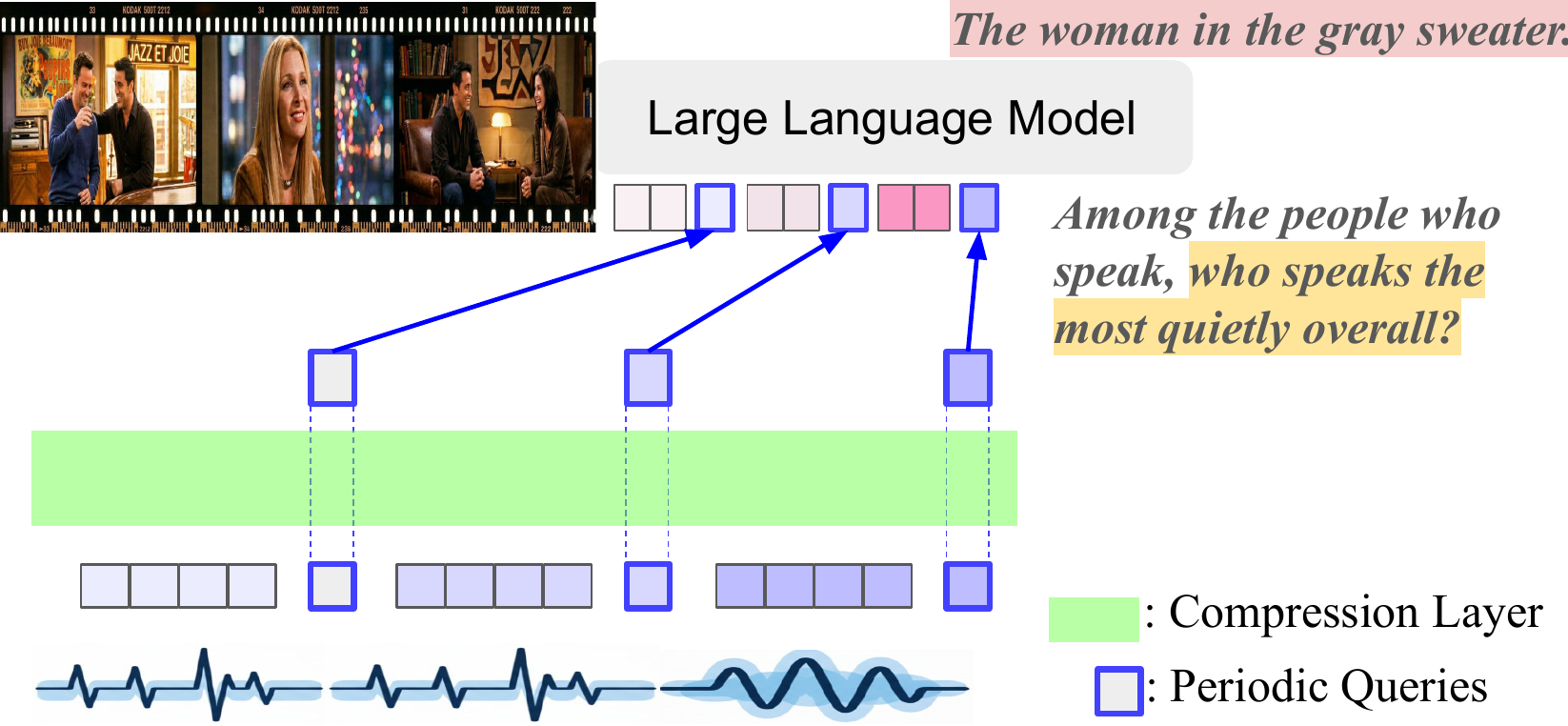}
    \caption{Mamba-based audio compressor with an AVSpeakerBench~\cite{avspeakerbench} example. The question (yellow) requires listening; the answer (red) requires watching.
    A periodic query every $R$ tokens yields $R{\times}$ reduction.}
    \label{fig:compressor}
\end{figure}

\subsection{Single-frame benchmark audit}
\label{subsec:audit}

To quantify how much current benchmarks rely on visual cues, we design a conservative single-frame probe. We feed only the temporally central frame---no audio, no other frames---to GPT-4o (\texttt{gpt-4o-2024-08-06}) in two independent runs at different temperatures; items answered correctly in both runs are removed. We additionally apply a refusal-aware prediction parser so that GPT-4o refusal responses (e.g., ``I cannot determine\ldots'') are not counted as correct, preventing false positives from coincidental letter matches in refusal text.

\begin{figure}[t]
    \centering
    \includegraphics[width=\linewidth]{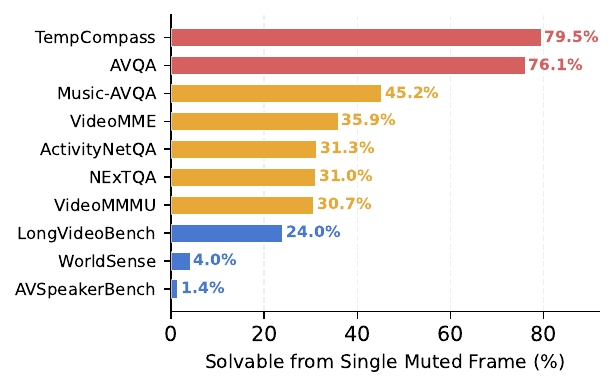}
    \caption{Fraction of items solvable from a single muted frame (GPT-4o, two runs at different temperatures, both correct). 
    }
    \label{fig:filter_rates}
\end{figure}

As shown in Fig.~\ref{fig:filter_rates}, filtering rates vary widely: ${\sim}$80\% of TempCompass and ${\sim}$76\% of AVQA items are answerable from a single muted frame, while WorldSense (4\%) and AVSpeakerBench (1\%) retain nearly all items.
This filtering is deliberately conservative: it removes only items solvable from a single muted frame, providing a lower bound on audio dependence. Comparing unfiltered and filtered halves of Table~\ref{tab:ablation} reveals which gaps are genuine vs.\ artifacts of visual shortcuts. 
We release the filtered splits to support fairer evaluation; details and qualitative examples are available in our repository.

\begin{table*}[t]
\centering
\caption{Audio integration ablation (top) and single-frame filtered evaluation (bottom). All audio rows use Avg Pool ($25\times$). Items solvable from a single muted frame (Fig.~\ref{fig:filter_rates}) are removed in the filtered subset. Results averaged over three random seeds. Small values under each benchmark are its approximate average\,/\,maximum video length.}
\label{tab:ablation}
\begin{adjustbox}{width=\linewidth}
\begin{tabular}{ccc|cccccccccc|c}
\toprule
\textbf{Audio} & \textbf{Intlv.} & \textbf{Comp.}
& \shortstack[c]{\textbf{AVQA}\\{\footnotesize 2m\,/\,45m}}
& \shortstack[c]{\textbf{AVSpkr-}\\\textbf{Bench}\\{\footnotesize 14s\,/\,66s}}
& \shortstack[c]{\textbf{Activity-}\\\textbf{NetQA}\\{\footnotesize 2m\,/\,16m}}
& \shortstack[c]{\textbf{LVideo-}\\\textbf{Bench}\\{\footnotesize 12m\,/\,1h}}
& \shortstack[c]{\textbf{Music-}\\\textbf{AVQA}\\{\footnotesize 1m\,/\,1m}}
& \shortstack[c]{\textbf{NExT-}\\\textbf{QA}\\{\footnotesize 40s\,/\,3m}}
& \shortstack[c]{\textbf{Temp-}\\\textbf{Comp.}\\{\footnotesize 11s\,/\,35s}}
& \shortstack[c]{\textbf{Video-}\\\textbf{MMMU}\\{\footnotesize 4m\,/\,34m}}
& \shortstack[c]{\textbf{Video-}\\\textbf{MME}\\{\footnotesize 17m\,/\,1h}}
& \shortstack[c]{\textbf{World-}\\\textbf{Sense}\\{\footnotesize 2m\,/\,11m}}
& \textbf{Avg} \\
\midrule
 & & -- & 90.8\pms{0.2} & 41.1\pms{0.9} & \textbf{53.0}\pms{0.5} & 54.5\pms{0.4} & \textbf{81.3}\pms{0.2} & 81.4\pms{0.2} & \textbf{63.7}\pms{0.6} & 32.7\pms{1.0} & 59.2\pms{0.5} & 39.3\pms{0.8} & 59.7 \\
\checkmark & & Avg Pool & 91.1\pms{0.6} & 42.5\pms{0.6} & 52.8\pms{0.3} & 55.3\pms{0.3} & 80.9\pms{0.1} & 81.5\pms{0.3} & 63.5\pms{0.8} & 33.7\pms{1.1} & 59.9\pms{1.4} & 40.9\pms{1.2} & 60.2 \\
\rowcolor{green!10}
\checkmark & \checkmark & Avg Pool & \textbf{91.6}\pms{0.3} & \textbf{44.1}\pms{0.7} & 52.6\pms{0.7} & \textbf{56.2}\pms{0.7} & 80.7\pms{0.1} & \textbf{81.6}\pms{0.4} & 63.5\pms{0.5} & \textbf{34.0}\pms{0.2} & \textbf{61.6}\pms{1.4} & \textbf{42.0}\pms{1.2} & \textbf{60.8} \\
\midrule
\midrule
\multicolumn{14}{l}{\textit{Results on single-frame filtered subset}} \\
 & & -- & 72.5\pms{0.1} & 40.7\pms{0.8} & \textbf{41.4}\pms{0.6} & 46.7\pms{0.3} & \textbf{72.0}\pms{0.2} & 76.0\pms{0.3} & \textbf{63.7}\pms{2.0} & 26.9\pms{1.1} & 49.6\pms{0.6} & 38.3\pms{0.8} & 52.8 \\
\rowcolor{green!10}
\checkmark & \checkmark & Avg Pool & \textbf{73.9}\pms{0.6} & \textbf{43.7}\pms{0.7} & 41.0\pms{0.5} & \textbf{48.6}\pms{0.9} & 71.1\pms{0.1} & \textbf{76.2}\pms{0.4} & 61.7\pms{0.5} & \textbf{27.6}\pms{0.5} & \textbf{52.0}\pms{1.6} & \textbf{41.0}\pms{1.2} & \textbf{53.7} \\
\bottomrule
\end{tabular}
\end{adjustbox}
\end{table*}

\subsection{Audio-visual input construction}
\label{subsec:arch_overview}

We build on the LLaVA family~\cite{liu2023llava,zhang2024llavanextvideo,li2025llavaonevision,zhang2025llavavideo}: a vision encoder (SigLIP2, 576 tokens per $384{\times}384$ frame) is aligned to an LLM (Qwen2-7B) via a lightweight projector. Let $\mathbf{V}$ and $\mathbf{A}$ denote visual and audio token sequences, respectively. We compare three input policies (Fig.~\ref{fig:overview}): (i)~visual-only ($\mathbf{V}$ alone), (ii)~non-interleaving ($[\mathbf{V};\,\mathbf{A}]$, all visual tokens first), and (iii)~time-aligned interleaving (audio tokens placed adjacent to their temporally corresponding frame tokens).
The audio encoder is the Whisper-based encoder of Qwen2-Audio~\cite{chu2024qwen2audiotechnicalreport} (the encoder only, not the full audio LLM); raw waveforms are converted to log-Mel spectrograms and processed by a Transformer stack with average pooling to 25\,Hz. Even at this rate, a 60-minute video yields ${\sim}$90K audio tokens, motivating the compression described next.

\subsection{Audio token compression}
\label{subsec:audio_compression}

We compress audio tokens by a factor of $R$ using a lightweight module inserted between the audio encoder and the LLM (Fig.~\ref{fig:compressor}).
The core idea is a \emph{periodic-query} design: given audio features $\mathbf{x}_{1:N}$ and stride $R$, we insert a shared trainable query $q$ every $R$ steps and pass the augmented sequence through a two-layer compression network. Only the outputs at query positions are retained:
\begin{align}
\label{eq:interleave}
\tilde{\mathbf{x}} &= \big[x_1,\ldots,x_{R-1},\, q,\, x_R,\ldots,x_{2R-1},\, q,\, \ldots \big],\\
\label{eq:select}
z_t &= \tilde{y}_{j_t}, \quad j_t \in \{ \text{positions of } q \text{ in } \tilde{\mathbf{x}} \},
\end{align}
where $\tilde{\mathbf{y}}$ is the network output. Although $q$ is shared, each $z_t$ absorbs different context. This yields $R$-fold reduction to $\lfloor N/R \rfloor$ tokens aligned to wall-clock time.
For a 60-minute video, $R{=}25$ reduces ${\sim}$90K tokens to ${\sim}$3.6K (1 token/s).
We instantiate the compression network with several architectures (Sec.~\ref{subsec:setup}); the comparison reveals which designs best preserve information for the 1-D, causal audio stream.

\begin{table}[t]
\centering
\caption{Compressor comparison on single-frame filtered subset scores (interleaved, $25\times$, averaged over three seeds).}
\label{tab:compressor}
\begin{adjustbox}{width=\linewidth}
\begin{tabular}{l|cccccc|c}
\toprule
\textbf{Compressor}
& \textbf{AVQA}
& \shortstack[c]{\textbf{AVSpkr-}\\\textbf{Bench}}
& \shortstack[c]{\textbf{LVideo-}\\\textbf{Bench}}
& \shortstack[c]{\textbf{Video-}\\\textbf{MMMU}}
& \shortstack[c]{\textbf{Video-}\\\textbf{MME}}
& \shortstack[c]{\textbf{World-}\\\textbf{Sense}}
& \textbf{Avg} \\
\midrule
Avg Pool & 73.9 & 43.7 & 48.6 & 27.6 & 52.0 & 41.0 & 47.8 \\
\midrule
Resampler~\cite{blip2} & 74.0 & 41.4 & 47.2 & 27.3 & 50.1 & 40.4 & 46.7 \\
\midrule
UniMamba~\cite{mamba2} & 73.6 & 43.9 & 48.3 & 27.2 & 52.2 & 42.5 & 48.0 \\
BiMamba~\cite{bimba} & 74.0 & 43.9 & 48.4 & \textbf{28.6} & 51.2 & 41.1 & 47.9 \\
\rowcolor{green!10}
UniMambaMia & \textbf{74.5} & \textbf{44.1} & \textbf{48.9} & 27.9 & \textbf{54.4} & \textbf{42.8} & \textbf{48.8} \\
\bottomrule
\end{tabular}
\end{adjustbox}
\end{table}

\begin{table*}[t]
\caption{Comparison with modern Video-LLMs. $^\dagger$Reproduced under unified evaluation; each cell shows original\,/\,filtered, where the filtered score is evaluated only on items \emph{not} answerable from a single muted frame (GPT-4o probe; Sec.~\ref{subsec:audit}). Latency is the mean per-sample wall time on VideoMME (single A100 80GB). Bold = best among Qwen2-7B models above the double line; \underline{underline} = best from Qwen2.5-Omni (different backbone and training scale). Rows below the double line use different LLM backbones or paper-reported numbers.}
\label{tab:sota}
\begin{adjustbox}{width=\linewidth}
\centering
\begin{tabular}{lc cc|cccccccccc|c}
    \toprule
    \textbf{Model}
    & \textbf{LLM}
    & \shortstack[c]{\textbf{Use}\\\textbf{Aud.}}
    & \shortstack[c]{\textbf{Lat.}\\\textbf{(s)}}
    & \textbf{AVQA}
    & \shortstack[c]{\textbf{AV-}\\\textbf{Spkr}}
    & \shortstack[c]{\textbf{Act-}\\\textbf{Net}}
    & \textbf{LVB}
    & \shortstack[c]{\textbf{M-}\\\textbf{AVQA}}
    & \shortstack[c]{\textbf{NExT-}\\\textbf{QA}}
    & \shortstack[c]{\textbf{Temp-}\\\textbf{C.}}
    & \shortstack[c]{\textbf{V-}\\\textbf{MMMU}}
    & \shortstack[c]{\textbf{Vid-}\\\textbf{MME}}
    & \textbf{WS}
    & \textbf{Avg}
    \\
    \midrule
    $^\dagger$Qwen2-VL~\cite{qwen2vl} & Qwen2-7B & $\times$ & 1.63 & \fv{86.6}{59.3} & \fv{42.6}{42.2} & \fv{51.4}{39.6} & \fv{22.9}{22.9} & \fv{60.3}{45.4} & \fv{79.3}{73.4} & \fv{\textbf{70.1}}{\textbf{64.2}} & \fv{32.9}{26.3} & \fv{57.5}{45.8} & \fv{36.6}{35.7} & \fv{54.0}{45.5} \\
    $^\dagger$LLaVA-OneVision~\cite{li2025llavaonevision} & Qwen2-7B & $\times$ & 0.97 & \fv{84.7}{59.7} & \fv{41.8}{41.5} & \fv{54.7}{42.5} & \fv{56.5}{48.3} & \fv{60.4}{48.7} & \fv{78.8}{72.9} & \fv{65.1}{60.8} & \fv{33.1}{26.8} & \fv{57.6}{47.2} & \fv{38.0}{36.7} & \fv{57.1}{48.5} \\
    $^\dagger$LLaVA-Video~\cite{zhang2025llavavideo} & Qwen2-7B & $\times$ & 1.00 & \fv{85.0}{57.6} & \fv{43.5}{43.3} & \fv{59.1}{47.2} & \fv{\textbf{58.4}}{\textbf{50.1}} & \fv{60.2}{45.2} & \fv{\textbf{81.8}}{\textbf{76.1}} & \fv{67.0}{\textbf{64.2}} & \fv{\textbf{38.3}}{\textbf{31.9}} & \fv{62.2}{52.6} & \fv{39.7}{38.5} & \fv{59.5}{50.7} \\
    \rowcolor{green!10}
    \textbf{Ours} & Qwen2-7B & \checkmark & 1.60 & \fv{\textbf{89.4}}{\textbf{70.0}} & \fv{\textbf{46.6}}{\textbf{46.3}} & \fv{\textbf{59.9}}{\textbf{49.2}} & \fv{53.9}{46.6} & \fv{\textbf{79.5}}{\textbf{69.2}} & \fv{\textbf{81.8}}{76.0} & \fv{63.7}{58.6} & \fv{36.7}{30.8} & \fv{\textbf{65.6}}{\textbf{56.6}} & \fv{\textbf{44.7}}{\textbf{43.9}} & \fv{\textbf{62.2}}{\textbf{54.7}} \\
    \midrule
    \midrule
    $^\dagger$LLaVA-NeXT-Vid.~\cite{zhang2024llavanextvideo} & Vicuna-7B & $\times$ & 0.24 & \fv{79.8}{54.8} & \fv{27.4}{27.4} & \fv{49.2}{36.5} & \fv{43.9}{37.2} & \fv{54.6}{47.0} & \fv{53.6}{44.1} & \fv{51.0}{38.6} & \fv{16.3}{14.7} & \fv{33.8}{29.1} & \fv{27.6}{26.8} & \fv{43.7}{35.6} \\
    $^\dagger$Qwen2.5-Omni~\cite{xu2025qwen25omnitechnicalreport} & Qwen2.5-7B & \checkmark & 4.12 & \fv{\underline{90.0}}{68.3} & \fv{\underline{47.1}}{\underline{47.0}} & \fv{48.3}{38.5} & \fv{22.4}{21.6} & \fv{45.7}{32.4} & \fv{76.8}{70.1} & \fv{69.6}{\underline{67.9}} & \fv{\underline{48.8}}{\underline{38.5}} & \fv{\underline{66.5}}{56.0} & \fv{\underline{45.4}}{\underline{44.5}} & \fv{56.1}{48.5} \\
    VideoLLaMA2~\cite{cheng2024videollama2advancingspatialtemporal} & Mistral-7B & \checkmark & - & - & - & 50.2\,/\,- & - & 73.6\,/\,- & - & - & - & 46.6\,/\,- & - & - \\
    VAMBA~\cite{vamba} & Qwen2-7B & $\times$ & - & - & - & - & - & - & 78.1\,/\,- & - & - & 57.8\,/\,- & - & - \\
    \bottomrule
\end{tabular}
\end{adjustbox}
\end{table*}

\section{Experiments and analysis}
\label{sec:experiments}

\subsection{Setup}
\label{subsec:setup}

\paragraph*{Benchmarks.}
We evaluate on 10 benchmarks: vision-centric suites (VideoMME~\cite{fu2024video}, TempCompass~\cite{liu-etal-2024-tempcompass-compress}, ActivityNetQA~\cite{actqa}, NExTQA~\cite{xiao2021next}, LongVideoBench~\cite{wu2024longvideobench}), audio-visual QA (Music-AVQA~\cite{Li2022Learning}, AVQA~\cite{yang2022avqa}, the latter on a retrievable $\sim$9K subset documented in our repository), and recent audio-visual suites (AVSpeakerBench~\cite{avspeakerbench}, VideoMMMU~\cite{videommmu}, WorldSense~\cite{worldsense}).

\paragraph*{Compressor variants.}
We instantiate five architectures within the periodic-query framing (Sec.~\ref{subsec:audio_compression}; Fig.~\ref{fig:compressor}):
(i)~\textbf{Avg Pool}---parameter-free average pooling followed by an MLP projector (${\sim}$17M params);
(ii)~\textbf{Resampler}~\cite{blip2}---attention-based compression with learnable queries;
(iii)~\textbf{UniMamba}~\cite{mamba2}---causal (unidirectional) SSM;
(iv)~\textbf{BiMamba}~\cite{bimba}---bidirectional SSM, widely used for video tokens but applied here to inherently sequential audio; and
(v)~\textbf{UniMambaMia}---adapted from MambaMia~\cite{kim2026statespace}, which compresses video tokens with a BiMamba backbone plus a gated attention module that re-weights compressed tokens by their pooled context. We replace the bidirectional backbone with a causal Mamba, retaining the gated attention while ensuring causal processing---a prerequisite for streaming audio.
Learned compressors (ii--v) each have ${\sim}$130M parameters; all variants share identical input/output shapes for fair comparison.

\paragraph*{Training.}
We compare three input configurations (Sec.~\ref{subsec:arch_overview}) combined with five compressor architectures. Ablations (Tables~\ref{tab:ablation}--\ref{tab:compressor}) use stride $R{=}25$ ($25\times$; 1 audio token/s); after comparing compressors (Table~\ref{tab:compressor}), we select UniMambaMia for the final model (Table~\ref{tab:sota}). All experiments are averaged over three random seeds.
We first perform image-level instruction tuning following ELVA~\cite{kim-seo-2024-efficient-comp}; we then insert the audio compressor and perform module-only alignment on subsets of LLaVA-Video-Set~\cite{zhang2025llavavideo} and FineVideo-Set~\cite{finevideo} (update compressor only); finally, we unfreeze the LLM and conduct video instruction tuning.
Video instruction data is constructed from LLaVA-Video-Set~\cite{zhang2025llavavideo}, FineVideo-Set~\cite{finevideo}, Music-AVQA v2 train set~\cite{liu2024tackling}, AVSD train set~\cite{alamri2019audiovisual}, and AVQA train set~\cite{yang2022avqa}. Ablations use 188K samples; the final model uses 420K. All experiments use 32 frames at 1.0~fps.
Both encoders are kept frozen for efficiency; we use a learning rate of $1{\times}10^{-4}$ for module-only alignment and $2{\times}10^{-5}$ for LLM tuning.

\subsection{Results and discussion}
\label{sec:results}

\paragraph*{Does audio help, and does interleaving matter?}
Table~\ref{tab:ablation} reports ablations on 10 benchmarks (three seeds, all using Avg Pool $25\times$). Adding audio with interleaving improves six of ten benchmarks---AVSpeakerBench ($+$3.0), AVQA, LongVideoBench, VideoMMMU, VideoMME ($+$2.4), and WorldSense ($+$2.7)---with the largest gains on tasks requiring speech comprehension or cross-modal grounding. ActivityNetQA, Music-AVQA, NExTQA, and TempCompass show marginal or no changes, consistent with their predominantly vision-centric designs (Fig.~\ref{fig:filter_rates}). Both AVQA and Music-AVQA training splits are included in our instruction data, yet their outcomes diverge: audio improves AVQA ($+$1.4\,pp after filtering) while Music-AVQA shows no benefit. This asymmetry is consistent with Fig.~\ref{fig:filter_rates}---Music-AVQA admits 45\% single-frame shortcuts and centers on musical content largely outside our speech-oriented encoder's domain, leaving minimal headroom for audio.
This contrast shows that including training data alone does not guarantee audio utility: \emph{benchmark design} determines whether the model learns to listen or to shortcut.

Interleaving yields only modest accuracy gains over non-interleaving (Table~\ref{tab:ablation}). We adopt it because (i)~it preserves temporal co-occurrence between modalities, and (ii)~paired with a causal compressor, it is the only configuration compatible with streaming inference. The absence of a BiMamba advantage over UniMamba (Table~\ref{tab:compressor}) further supports a causal design.

\paragraph*{Do gains survive single-frame filtering?}
Filtering (Table~\ref{tab:ablation}, bottom) removes visually solvable items, sharpening the picture. 
AVQA scores drop from ${\sim}$92 to ${\sim}$73, confirming that many items were answerable from visual cues alone.
After filtering, audio produces clear gains on 5 of 10 benchmarks: AVSpeakerBench ($+$3.0\,pp), WorldSense ($+$2.7\,pp), VideoMME ($+$2.4\,pp), LongVideoBench ($+$1.9\,pp), and AVQA ($+$1.4\,pp)---all centered on speech comprehension or cross-modal grounding. Across VideoMME's three duration categories, the filtered gain widens from $+$0.6 (short) to $+$1.8 (medium) to $+$4.4 (long): with the frame budget fixed, longer videos are sampled more sparsely, so frames alone capture less and audio contributes more.
Benchmarks that decline slightly (ActivityNetQA, Music-AVQA, TempCompass) are predominantly vision-centric; for these, audio tokens carry little task-relevant signal and can mildly interfere. 
TempCompass shows a small filtered drop, but filtering removes ${\sim}$80\% of items, leaving only 324 questions; TempCompass is almost entirely vision-centric by design, so the surviving items still rarely require audio, and filtered scores on such a small set are inherently noisy (e.g., AVQA retains 2{,}194; VideoMME 1{,}731).
WorldSense---which retains 96\% of items after filtering---shows a gain of $+$2.7\,pp both before and after filtering, consistent with its already audio-demanding design.
This result is compressor-agnostic: even the simplest baseline (Avg Pool) reveals audio's value, and a better compressor (Table~\ref{tab:compressor}) can further amplify gains.

\paragraph*{Which compressor works best?}
Having established that audio helps even with a parameter-free compressor (Table~\ref{tab:ablation}), we now ask which architecture best preserves information under $25\times$ compression.
Table~\ref{tab:compressor} compares five compressors on filtered scores (interleaved, $25\times$).
The main findings are threefold.
First, Mamba-based compressors outperform Avg Pool, confirming that the token-compression techniques recently developed for vision~\cite{shen2025longvu,vamba,kim2026statespace} transfer effectively to audio.
Second, unlike the 2-D video setting where bidirectional models often dominate, BiMamba offers no clear advantage over causal UniMamba on any benchmark. This is consistent with the inherently sequential, 1-D nature of audio, where future context provides limited additional information.
Third, UniMambaMia achieves the best or tied-best scores on five of six benchmarks (AVQA, AVSpeakerBench, LongVideoBench, VideoMME, WorldSense), making it the most consistent choice.
Given the comparable accuracy of uni- and bidirectional designs, we prefer the causal variant because it is the only one compatible with streaming inference.

\begin{figure}[t!]
    \centering
    \includegraphics[width=\linewidth]{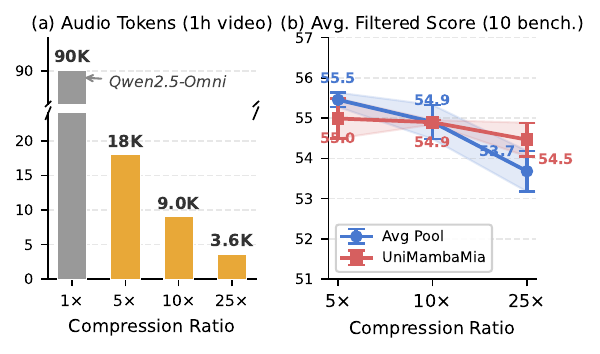}
    \caption{(a)~Audio token count for a one-hour video; without compression, 90K tokens are consumed (e.g., Qwen2.5-Omni). (b)~Avg.\ filtered score (10 benchmarks) vs.\ compression ratio. UniMambaMia degrades less at $25\times$ ($-$0.5\,pp vs.\ $-$1.8\,pp).}
    \label{fig:compression_ratio}
\end{figure}

\paragraph*{Compression ratio sweep.}
Fig.~\ref{fig:compression_ratio}(a) shows that $25\times$ compression reduces a one-hour video's 90K tokens to 3.6K. Panel~(b) confirms UniMambaMia degrades more gracefully than Avg Pool at $25\times$, motivating its adoption.

\paragraph*{How does our model compare with modern Video-LLMs?}
Based on Tables~\ref{tab:ablation}--\ref{tab:compressor}, we select UniMambaMia (interleaved, $25\times$) as our final model.
Under unified evaluation (Table~\ref{tab:sota}), it achieves the best or tied-best results on 7 of 10 benchmarks among Qwen2-7B models; filtered scores confirm these gains persist after removing visual shortcuts.
Differences here are confounded by backbone and training data: Qwen2.5-Omni (Qwen2.5-7B, proprietary data) leads on VideoMMMU, while our model leads on Music-AVQA (79.5 vs.\ 45.7). These gaps reflect more than the audio pathway alone.
Our model adds audio at moderate latency cost (1.60\,s vs.\ 1.00\,s for LLaVA-Video) by compressing to ${\sim}$3.6K tokens per hour; Qwen2.5-Omni feeds uncompressed audio (${\sim}$90K tokens/hour) at 4.12\,s and ${\sim}$62\,GiB peak inference-time GPU memory in our setup, versus ${\sim}$34\,GiB for ours, illustrating the practical value of compression.

\section{Conclusion}
\label{sec:conclusion}
Our experiments yield a clear answer to the title question: yes, modern Video-LLMs do need to listen---but only if benchmarks are designed to require it.
Many widely used suites admit vision-only shortcuts, hiding the true value of listening.
Once these shortcuts are controlled for, audio produces clear gains on speech- and audio-dependent tasks.
On the modeling side, time-aligned interleaving with a lightweight causal compressor offers a simple, scalable recipe that retains audio benefits while largely preserving visual performance.
The gap between muted evaluation and real-world use---where lectures, meetings, and everyday videos carry essential speech and sound---remains wide.
By showing that muting systematically underestimates the value of speech and audio representations, we aim to encourage more rigorous audio-visual evaluation and help close this gap between benchmarks and deployment.

\section{Generative AI Use Disclosure}
Generative AI tools (e.g., Claude) were used for editing and polishing the manuscript text. All scientific content, experimental design, analysis, and conclusions are the sole responsibility of the authors.

\bibliographystyle{IEEEtran}
\bibliography{refs}

\begin{thebibliography}{10}
\providecommand{\url}[1]{#1}
\csname url@samestyle\endcsname
\providecommand{\newblock}{\relax}
\providecommand{\bibinfo}[2]{#2}
\providecommand{\BIBentrySTDinterwordspacing}{\spaceskip=0pt\relax}
\providecommand{\BIBentryALTinterwordstretchfactor}{4}
\providecommand{\BIBentryALTinterwordspacing}{\spaceskip=\fontdimen2\font plus
\BIBentryALTinterwordstretchfactor\fontdimen3\font minus \fontdimen4\font\relax}
\providecommand{\BIBforeignlanguage}[2]{{%
\expandafter\ifx\csname l@#1\endcsname\relax
\typeout{** WARNING: IEEEtran.bst: No hyphenation pattern has been}%
\typeout{** loaded for the language `#1'. Using the pattern for}%
\typeout{** the default language instead.}%
\else
\language=\csname l@#1\endcsname
\fi
#2}}
\providecommand{\BIBdecl}{\relax}
\BIBdecl

\bibitem{whisper}
A.~Radford, J.~W. Kim, T.~Xu, G.~Brockman, C.~McLeavey, and I.~Sutskever, ``{Robust Speech Recognition via Large-Scale Weak Supervision},'' in \emph{ICML}, 2023.

\bibitem{liu2023llava}
H.~Liu, C.~Li, Q.~Wu, and Y.~J. Lee, ``{Visual Instruction Tuning},'' in \emph{NeurIPS}, 2023.

\bibitem{zhang2024llavanextvideo}
\BIBentryALTinterwordspacing
Y.~Zhang, B.~Li, H.~Liu, Y.~J. Lee, L.~Gui, D.~Fu, J.~Feng, Z.~Liu, and C.~Li, ``{LLaVA-NeXT: A Strong Zero-shot Video Understanding Model},'' April 2024. [Online]. Available: \url{https://llava-vl.github.io/blog/2024-04-30-llava-next-video/}
\BIBentrySTDinterwordspacing

\bibitem{li2025llavaonevision}
B.~Li, Y.~Zhang, D.~Guo, R.~Zhang, F.~Li, H.~Zhang, K.~Zhang, P.~Zhang, Y.~Li, Z.~Liu, and C.~Li, ``{LLaVA-OneVision: Easy Visual Task Transfer},'' \emph{TMLR}, 2025.

\bibitem{zhang2025llavavideo}
Y.~Zhang, J.~Wu, W.~Li, B.~Li, Z.~Ma, Z.~Liu, and C.~Li, ``{LLaVA-Video: Video Instruction Tuning With Synthetic Data},'' \emph{TMLR}, 2025.

\bibitem{qwen2vl}
\BIBentryALTinterwordspacing
P.~Wang, S.~Bai, S.~Tan, S.~Wang, Z.~Fan, J.~Bai, K.~Chen, X.~Liu, J.~Wang, W.~Ge, Y.~Fan, K.~Dang, M.~Du, X.~Ren, R.~Men, D.~Liu, C.~Zhou, J.~Zhou, and J.~Lin, ``{Qwen2-VL: Enhancing Vision-Language Model's Perception of the World at Any Resolution},'' 2024. [Online]. Available: \url{https://arxiv.org/abs/2409.12191}
\BIBentrySTDinterwordspacing

\bibitem{actqa}
Z.~Yu, D.~Xu, J.~Yu, T.~Yu, Z.~Zhao, Y.~Zhuang, and D.~Tao, ``{ActivityNet-QA: a dataset for understanding complex web videos via question answering},'' in \emph{AAAI}, 2019.

\bibitem{xiao2021next}
J.~Xiao, X.~Shang, A.~Yao, and T.-S. Chua, ``{NExT-QA: Next Phase of Question-Answering to Explaining Temporal Actions},'' in \emph{CVPR}, 2021.

\bibitem{liu-etal-2024-tempcompass-compress}
Y.~Liu, S.~Li, Y.~Liu, Y.~Wang, S.~Ren, L.~Li, S.~Chen, X.~Sun, and L.~Hou, ``{TempCompass: Do Video LLMs Really Understand Videos?}'' in \emph{Findings of ACL}, 2024.

\bibitem{yang2022avqa}
P.~Yang, X.~Wang, X.~Duan, H.~Chen, R.~Hou, C.~Jin, and W.~Zhu, ``{AVQA: A Dataset for Audio-Visual Question Answering on Videos},'' in \emph{ACM Multimedia}, 2022.

\bibitem{sung2024avhbench}
S.-B. Kim, H.-B. Oh, J.~Lee, A.~Senocak, J.~S. Chung, and T.-H. Oh, ``{AVHBench}: A cross-modal hallucination benchmark for audio-visual large language models,'' in \emph{ICLR}, 2025.

\bibitem{worldsense}
J.~Hong, S.~Yan, J.~Cai, X.~Jiang, Y.~Hu, and W.~Xie, ``{WorldSense: Evaluating Real-world Omnimodal Understanding for Multimodal LLMs},'' in \emph{ICLR}, 2026.

\bibitem{avspeakerbench}
L.~T.~P. Nguyen, Z.~Yu, S.~L.~Y. Hang, S.~An, J.~Lee, Y.~Ban, S.~Chung, T.-H. Nguyen, J.~Maeng, S.~Lee, and Y.~J. Lee, ``{See, Hear, and Understand: Benchmarking Audiovisual Human Speech Understanding in Multimodal Large Language Models},'' in \emph{CVPR Findings}, 2026.

\bibitem{avut}
Y.~Yang, J.~Zhuang, G.~Sun, C.~Tang, Y.~Li, P.~Li, Y.~Jiang, W.~Li, Z.~Ma, and C.~Zhang, ``{Audio-centric Video Understanding Benchmark without Text Shortcut},'' in \emph{EMNLP}, 2025.

\bibitem{fu2024video}
C.~Fu, Y.~Dai, Y.~Luo, L.~Li, S.~Ren, R.~Zhang, Z.~Wang, C.~Zhou, Y.~Shen, M.~Zhang, P.~Chen, Y.~Li, S.~Lin, S.~Zhao, K.~Li, T.~Xu, X.~Zheng, E.~Chen, C.~Shan, R.~He, and X.~Sun, ``{Video-MME: The First-Ever Comprehensive Evaluation Benchmark of Multi-modal LLMs in Video Analysis},'' in \emph{CVPR}, 2025.

\bibitem{wu2024longvideobench}
H.~Wu, D.~Li, B.~Chen, and J.~Li, ``{LongVideoBench: A Benchmark for Long-context Interleaved Video-Language Understanding},'' in \emph{NeurIPS Datasets and Benchmarks Track}, 2024.

\bibitem{xu2025qwen25omnitechnicalreport}
\BIBentryALTinterwordspacing
J.~Xu, Z.~Guo, J.~He, H.~Hu, T.~He, S.~Bai, K.~Chen, J.~Wang, Y.~Fan, K.~Dang, B.~Zhang, X.~Wang, Y.~Chu, and J.~Lin, ``{Qwen2.5-Omni Technical Report},'' 2025. [Online]. Available: \url{https://arxiv.org/abs/2503.20215}
\BIBentrySTDinterwordspacing

\bibitem{cheng2024videollama2advancingspatialtemporal}
\BIBentryALTinterwordspacing
Z.~Cheng, S.~Leng, H.~Zhang, Y.~Xin, X.~Li, G.~Chen, Y.~Zhu, W.~Zhang, Z.~Luo, D.~Zhao, and L.~Bing, ``{VideoLLaMA 2: Advancing Spatial-Temporal Modeling and Audio Understanding in Video-LLMs},'' 2024. [Online]. Available: \url{https://arxiv.org/abs/2406.07476}
\BIBentrySTDinterwordspacing

\bibitem{blip2}
J.~Li, D.~Li, S.~Savarese, and S.~Hoi, ``{BLIP-2: Bootstrapping Language-Image Pre-training with Frozen Image Encoders and Large Language Models},'' in \emph{ICML}, 2023.

\bibitem{Li2022Learning}
G.~Li, Y.~Wei, Y.~Tian, C.~Xu, J.-R. Wen, and D.~Hu, ``{Learning To Answer Questions in Dynamic Audio-Visual Scenarios},'' in \emph{CVPR}, 2022.

\bibitem{gong2024avodyssey}
\BIBentryALTinterwordspacing
K.~Gong, K.~Feng, B.~Li, Y.~Wang, M.~Cheng, S.~Yang, J.~Han, B.~Wang, Y.~Bai, Z.~Yang, and X.~Yue, ``{AV-Odyssey Bench: Can Your Multimodal LLMs Really Understand Audio-Visual Information?}'' 2024. [Online]. Available: \url{https://arxiv.org/abs/2412.02611}
\BIBentrySTDinterwordspacing

\bibitem{li2024omnibench}
Y.~Li, Y.~Ma, G.~Zhang, R.~Yuan, K.~Zhu, H.~Guo, Y.~Liang, J.~Liu, Z.~Wang, J.~Yang, S.~Wu, X.~Qu, J.~Shi, X.~Zhang, Z.~Yang, Y.~Wen, Y.~Wang, S.~Li, Z.~Zhang, R.~Liu, E.~Benetos, W.~Huang, and C.~Lin, ``{OmniBench: Towards The Future of Universal Omni-Language Models},'' in \emph{NeurIPS Datasets and Benchmarks Track}, 2025.

\bibitem{chen2025savvy}
M.~Chen, Z.~Cui, X.~Liu, J.~Xiang, C.~Zheng, J.~Li, and E.~Shlizerman, ``{SAVVY: Spatial Awareness via Audio-Visual {LLM}s through Seeing and Hearing},'' in \emph{NeurIPS}, 2025.

\bibitem{shen2025longvu}
X.~Shen, Y.~Xiong, C.~Zhao, L.~Wu, J.~Chen, C.~Zhu, Z.~Liu, F.~Xiao, B.~Varadarajan, F.~Bordes, Z.~Liu, H.~Xu, H.~J. Kim, B.~Soran, R.~Krishnamoorthi, M.~Elhoseiny, and V.~Chandra, ``{LongVU: Spatiotemporal Adaptive Compression for Long Video-Language Understanding},'' in \emph{ICML}, 2025.

\bibitem{vamba}
W.~Ren, W.~Ma, H.~Yang, C.~Wei, G.~Zhang, and W.~Chen, ``{Vamba: Understanding Hour-Long Videos with Hybrid Mamba-Transformers},'' in \emph{ICCV}, 2025.

\bibitem{videomamba}
K.~Li, X.~Li, Y.~Wang, Y.~He, Y.~Wang, L.~Wang, and Y.~Qiao, ``{VideoMamba: State Space Model for Efficient Video Understanding},'' in \emph{ECCV}, 2024.

\bibitem{kim2026statespace}
G.~Kim and M.~Seo, ``{State-Space Hierarchical Compression with Gated Attention and Learnable Sampling for Hour-Long Video Understanding in Large Multimodal Models},'' in \emph{AAAI}, 2026.

\bibitem{chu2024qwen2audiotechnicalreport}
\BIBentryALTinterwordspacing
Y.~Chu, J.~Xu, Q.~Yang, H.~Wei, X.~Wei, Z.~Guo, Y.~Leng, Y.~Lv, J.~He, J.~Lin, C.~Zhou, and J.~Zhou, ``{Qwen2-Audio Technical Report},'' 2024. [Online]. Available: \url{https://arxiv.org/abs/2407.10759}
\BIBentrySTDinterwordspacing

\bibitem{mamba2}
T.~Dao and A.~Gu, ``{Transformers are SSMs: Generalized Models and Efficient Algorithms Through Structured State Space Duality},'' in \emph{ICML}, 2024.

\bibitem{bimba}
M.~M. Islam, T.~Nagarajan, H.~Wang, G.~Bertasius, and L.~Torresani, ``{BIMBA: Selective-Scan Compression for Long-Range Video Question Answering},'' in \emph{CVPR}, 2025.

\bibitem{videommmu}
\BIBentryALTinterwordspacing
K.~Hu, P.~Wu, F.~Pu, W.~Xiao, Y.~Zhang, X.~Yue, B.~Li, and Z.~Liu, ``{Video-MMMU: Evaluating Knowledge Acquisition from Multi-Discipline Professional Videos},'' 2025. [Online]. Available: \url{https://arxiv.org/abs/2501.13826}
\BIBentrySTDinterwordspacing

\bibitem{kim-seo-2024-efficient-comp}
G.~Kim and M.~Seo, ``{On Efficient Language and Vision Assistants for Visually-Situated Natural Language Understanding: What Matters in Reading and Reasoning},'' in \emph{EMNLP}, 2024.

\bibitem{finevideo}
\BIBentryALTinterwordspacing
M.~Farré, A.~Marafioti, L.~Tunstall, L.~Von~Werra, and T.~Wolf, ``{FineVideo},'' 2024. [Online]. Available: \url{https://huggingface.co/datasets/HuggingFaceFV/finevideo}
\BIBentrySTDinterwordspacing

\bibitem{liu2024tackling}
X.~Liu, Z.~Dong, and P.~Zhang, ``{Tackling Data Bias in MUSIC-AVQA: Crafting a Balanced Dataset for Unbiased Question-Answering},'' in \emph{WACV}, 2024.

\bibitem{alamri2019audiovisual}
H.~Alamri, V.~Cartillier, A.~Das, J.~Wang, A.~Cherian, I.~Essa, D.~Batra, T.~K. Marks, C.~Hori, P.~Anderson, S.~Lee, and D.~Parikh, ``{Audio Visual Scene-Aware Dialog},'' in \emph{CVPR}, June 2019.

\end{thebibliography}

\end{document}